\documentclass[letterpaper, 10 pt, conference]{ieeeconf}  
\usepackage{times}
\usepackage{booktabs}
\usepackage[backend=biber]{biblatex}
\usepackage{multicol}
\usepackage[utf8]{inputenc}
\usepackage{graphicx}
\usepackage{lipsum}
\usepackage{wrapfig}
\usepackage{hyperref}
\usepackage{algpseudocode}
\usepackage{algorithm}
\usepackage{soul}
\usepackage{todonotes}
\usepackage{amsmath, amssymb}
\usepackage{cleveref}
\usepackage{bm}
\usepackage{color}
\usepackage{multirow}
\usepackage{comment}
\usepackage{float}
\usepackage{caption}
\usepackage{subfigure}
\usepackage{siunitx}
\usepackage{float}
\usepackage{adjustbox}
\usepackage{graphicx}
\usepackage{hyperref}
\definecolor{Burgundy1}{RGB}{128,0,32}
\def\imagetop#1{\vtop{\null\hbox{#1}}}

\let\oldnl\nl
\newcommand{\nonl}{\renewcommand{\nl}{\let\nl\oldnl}}
\algnewcommand\algorithmicforeach{\textbf{for each}}
\algdef{S}[FOR]{ForEach}[1]{\algorithmicforeach\ #1\ \algorithmicdo}
\newcommand{\cmmnt}[1]{}

\DeclareMathOperator*{\argmin}{arg\,min}

\IEEEoverridecommandlockouts                              

\title{\LARGE \bf
Optimizing Trajectories for Highway Driving with Offline Reinforcement Learning
}

\author{Branka Mirchevska$^{1}$, Moritz Werling$^{2}$, Joschka Boedecker$^{1, 3}$
\thanks{$^{1}$Dept. of Computer Science, University of Freiburg, Germany.}%
\thanks{{\tt \{mirchevb,jboedeck\}@cs.uni-freiburg.de}}
\thanks{$^{2}$BMW Group, Unterschleissheim, Germany.}%
\thanks{{\tt  Moritz.Werling@bmw.de}}%
\thanks{$^{3}$Cluster of Excellence BrainLinks-BrainTools, Freiburg, Germany.}%
}

\begin{document}

\maketitle
\thispagestyle{empty}
\pagestyle{empty}

\begin{abstract}\label{abstract}
Implementing an autonomous vehicle that is able to output feasible, smooth and efficient trajectories is a long-standing challenge.
Several approaches have been considered, roughly falling under two categories: rule-based and learning-based approaches.
The rule-based approaches, while guaranteeing safety and feasibility, fall short when it comes to long-term planning and generalization.
The learning-based approaches are able to account for long-term planning and generalization to unseen situations, but may fail to achieve smoothness, safety and the feasibility which rule-based approaches ensure.
Hence, combining the two approaches is an evident step towards yielding the best compromise out of both. 
We propose a Reinforcement Learning-based approach, which learns target trajectory parameters for fully autonomous driving on highways.
The trained agent outputs continuous trajectory parameters based on which a feasible polynomial-based trajectory is generated and executed.
We compare the performance of our agent 
against four other highway driving agents.
The experiments are conducted in the Sumo simulator, taking into consideration various realistic, dynamically changing highway scenarios, including surrounding vehicles with different driver behaviors.
We demonstrate that our offline trained agent, with randomly collected data, learns to drive smoothly, achieving velocities as close as possible to the desired velocity, while outperforming the other agents.
Code, training data and details available at: \url{https://nrgit.informatik.uni-freiburg.de/branka.mirchevska/offline-rl-tp}.
\end{abstract}

\section{Introduction}
In the recent past, autonomous driving (AD) on highways has been a very active area of research \cite{KATRAKAZAS2015416}, \cite{Schwarting2018PlanningAD}.
Numerous approaches have been proposed that can be broadly divided into two categories:
approaches that do not learn from data such as rule-based or optimal-control based approaches \cite{Claussmann}, and machine learning-based (ML) approaches \cite{2020Survey}. 
The former rely either on a set of rules tuned by human experience, or are formulated as an optimization problem subject to various constraints, aiming for a solution with the lowest cost \cite{5509799,6225063,5462899,Borrelli2005MPCBasedAT,Falcone2007PredictiveAS}.
Usually they are based on mathematically sound safety rules and are able to provide feasible and smooth driving trajectories \cite{articleRaoAnil}. 
However, it is increasingly difficult to define rules general enough to account for every situation that may occur, while ensuring the absence of errors caused by false human assumptions. 
Also, optimization-based approaches can get computationally intensive in complex, highly dynamic traffic settings due to the real-time constraint of the solver. 
Additionally, the limited planning horizon in optimization based approaches results in the inability for long-term planning, which may lead to a sub-optimal overall performance. 
Alternatively, the ML-based approaches learn directly from data and can generalize well over unseen situations \cite{8951131}. 
Since acquiring high-quality labeled data is cumbersome and expensive, Reinforcement Learning (RL) in particular offers a good alternative for tackling the autonomous driving problem \cite{Mirchevska2018HighlevelDM,Hgle2019DynamicIF,DBLP:journals/corr/abs-2012-03234,wang2019deep,8914621}.
Based on the type of the actions the agent learns to perform, we can divide the RL-based approaches into high-level, low-level, and approaches combining the two. 
High-level action approaches are the ones where the agent may choose from a few discrete actions defining some high-level maneuver like keep lane or perform a lane-change \cite{Mirchevska2018HighlevelDM,Hgle2019DynamicIF,branka_rl_highway}. 
In low-level approaches such as \cite{wang2019deep,wang2018reinforcement,wang2019quadratic} the agent usually outputs actions that directly influence the lowest level of control such as acceleration and steering wheel. 
Since the low-level approaches do not rely on any underlying structure, everything needs to be learned from scratch, requiring large amounts of high-quality data. 
At the same time, high-level approaches may be too limited since the maneuvers on the lower level are usually hard-coded and fixed in terms of duration and shape.

\begin{figure}[t]
	\centering
	\includegraphics[scale=0.3]{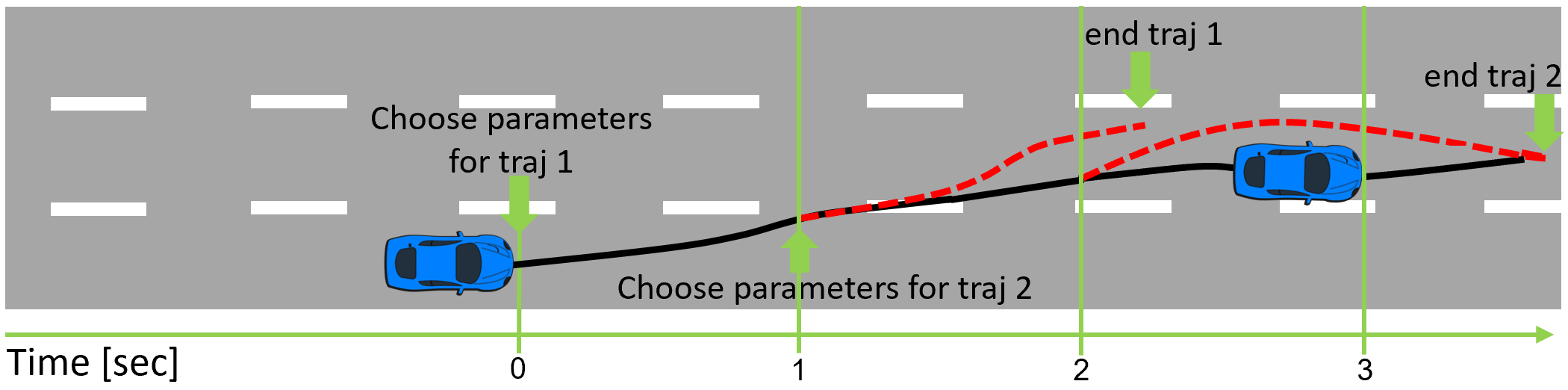}
	\caption{Our approach combines Reinforcement Learning and a polynomial-based trajectory planner to generate updated trajectories every second. The RL algorithm sets the desired target values for parameters of the trajectories including lateral position, lateral profile duration, longitudinal velocity and longitudinal profile duration.}
	\label{traj_plan}
\end{figure} 

To get the best of both worlds, the combination of RL and a model-based method holds the promise to speed up and improve the learning process. 
It enables a balance between high- and low-level actions by allowing enough flexibility for the RL agent while avoiding the need to learn tasks that are simple to handle by traditional control methods.
Most methods combining trajectory optimization with RL \cite{DBLP:journals/corr/abs-2107-06629,8968010,DBLP:journals/corr/abs-1910-09667,DBLP:journals/corr/abs-2012-03234}, separate the RL agent from the trajectory decision.
In \cite{DBLP:journals/corr/abs-2012-03234} we developed an RL agent that first chooses the best candidate from a set of gaps (space between two surrounding vehicles on the same lane) to fit into, and then triggers a trajectory planner to choose the best trajectory to get there.
Even though the agent performed well, this procedure is costly since it entails defining all gaps and the trajectories reaching them at run-time, including checks for collisions and assessing velocities before proposing the available actions to the agent.

To address this, here we are proposing a continuous control trajectory optimization approach based on RL.
Every second, the agent chooses four continuous actions describing the target trajectory parameters as shown in Fig.~ \ref{traj_plan}.
This provides the flexibility to directly choose the parameters for the generation of the trajectory that it is going to be executed until the next time-step.
Furthermore, instead of learning how to generate the trajectories, we delegate this task to a well-established polynomial-based  trajectory generation module.

In many applications, the need for learning a good policy using only a pre-collected set of data \footnote{Known as offline or fixed-batch RL}, becomes evident since learning while interacting with the environment is either too expensive or dangerous.
Autonomous driving is considered to be one of them, where in production application, the data is usually first collected and then used offline.
Considering that, we conduct the training using only data collected with a simple random policy in an offline fashion.
We evaluate our trained agent on realistic highly dynamic highway scenarios and show that it learned to drive as close as possible to the desired velocity, while outperforming the comparison agents.
It is also able to handle unpredictable situations such as sudden cut-ins and scenarios where more complex maneuvers are required. 
Additionally, we inspect how the nature of the data and the percentage of terminal samples influence the learned policy. 
Our main contributions are the following:
\begin{enumerate}
	\item Novel offline RL-based approach for AD on highways with continuous control for separate lateral and longitudinal planning components resting on an underlying
	polynomial-based trajectory generation module.
	\item Comparison against other models, on diverse realistic scenarios with surrounding vehicles controlled by different driver types.
	\item Demonstrating that the agent has learned to successfully deal with critical situations such as sudden cut-ins and scenarios where more complex maneuvers are required.
	\item Training data analysis based on the structure of the data, as well as the portion of terminal samples.
\end{enumerate}

\section{Reinforcement Learning Background}\label{rl}
In RL an agent learns to perform specific tasks from interactions with the environment. 
RL problems are usually modeled as a Markov Decision Process (MDP) $\langle \mathcal{S, A, T}, r, \gamma \rangle$, where $\mathcal{S} \in \mathbb{R}^n$ is the set of states, $\mathcal{A} \in \mathbb{R}^m$ the set of actions and $\mathcal{T}: \mathcal{S} \times \mathcal{A} \rightarrow \mathcal{S}$ is the transition model.
At each time-step $t$, the agent finds itself in a state $s_t$ and acts by executing an action $a_t$, in the environment. 
As a consequence of taking action $a_t$, the agent transitions to the next state $s_{t+1}$ and receives a feedback value $r_{t+1}$ from the environment, called a reward. 
The feedback is a measure of success or failure of the agent’s actions. 
The agent’s objective is to find a policy $\pi$, i.e., a mapping from states to actions that maximizes the expected discounted sum of rewards accumulated over time, defined as $\mathbb{E_\pi} \sum_{t=0}^{\infty} \gamma^t r_{t+1}$.
The discount factor $\gamma \in [0, 1]$ regulates the importance of future rewards.
The performance of the agent, following the policy $\pi$, can be measured using the action-value function $Q^\pi(s, a) = \mathbb{E}[\sum_{t=0}^{\infty}\gamma^t r_{t+1}|s_0\!=s, a_0\!=a]$.
The agent infers the optimal policy at a state $s_t$ from the optimal state-action value function $Q^*(s_t, a_t) = \max_{\pi}Q^\pi(s, a)$, via maximization. 
Another way to learn the parameterized policy is directly by using policy gradient methods.

The common way of dealing with RL problems with continuous actions are the \textbf{Actor-Critic methods}, which are integrating both value iteration and policy gradient.
The actor represents the policy and is used to select actions, whereas the critic represents the estimated value function and criticizes the actions taken by the actor.

In \textbf{offline RL}, instead of learning while interacting with the environment, the agent is provided with a fixed batch of transition samples, i.e., interactions with the environment, to learn from.
This data has been previously collected by one or a few policies unknown to the agent.
Offline RL is considered more challenging than online RL because the agent does not get to explore the environment based on the current policy, but it is expected to learn only from pre-collected data.
However, offline learning is beneficial for problems for which acquiring new data is impossible, or interaction with the environment is expensive or dangerous.
For more details and background on RL refer to \cite{Sutton1998}.

\section{Approach}\label{method}

\begin{figure}[t]
	\centering
	\includegraphics[scale=0.29]{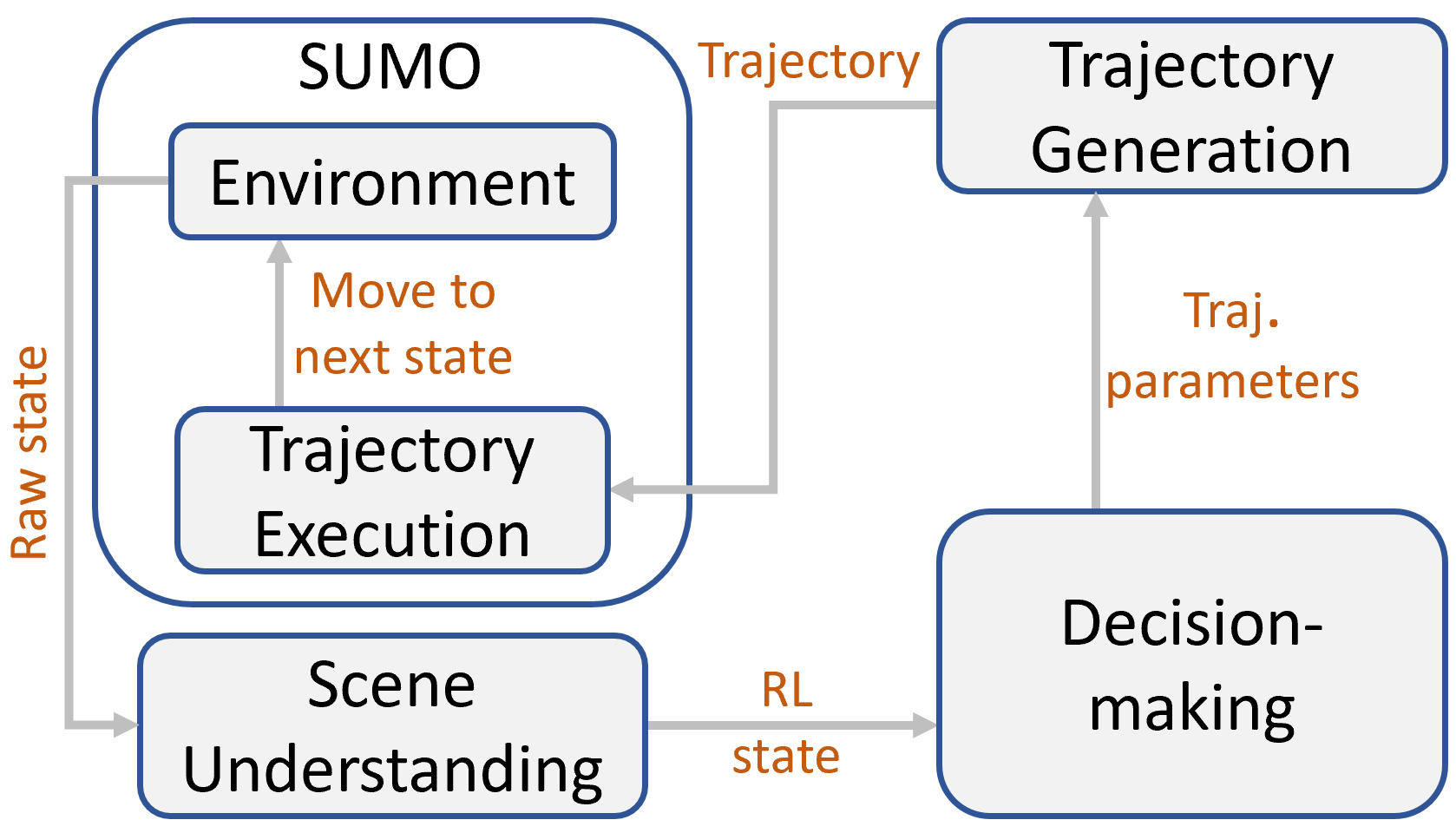}
	\caption{Approach modules}
	\label{approach_modules}
\end{figure}

\begin{figure}[t]
	\centering
	\hspace*{-0.2cm}\includegraphics[width=7cm, height=6cm]{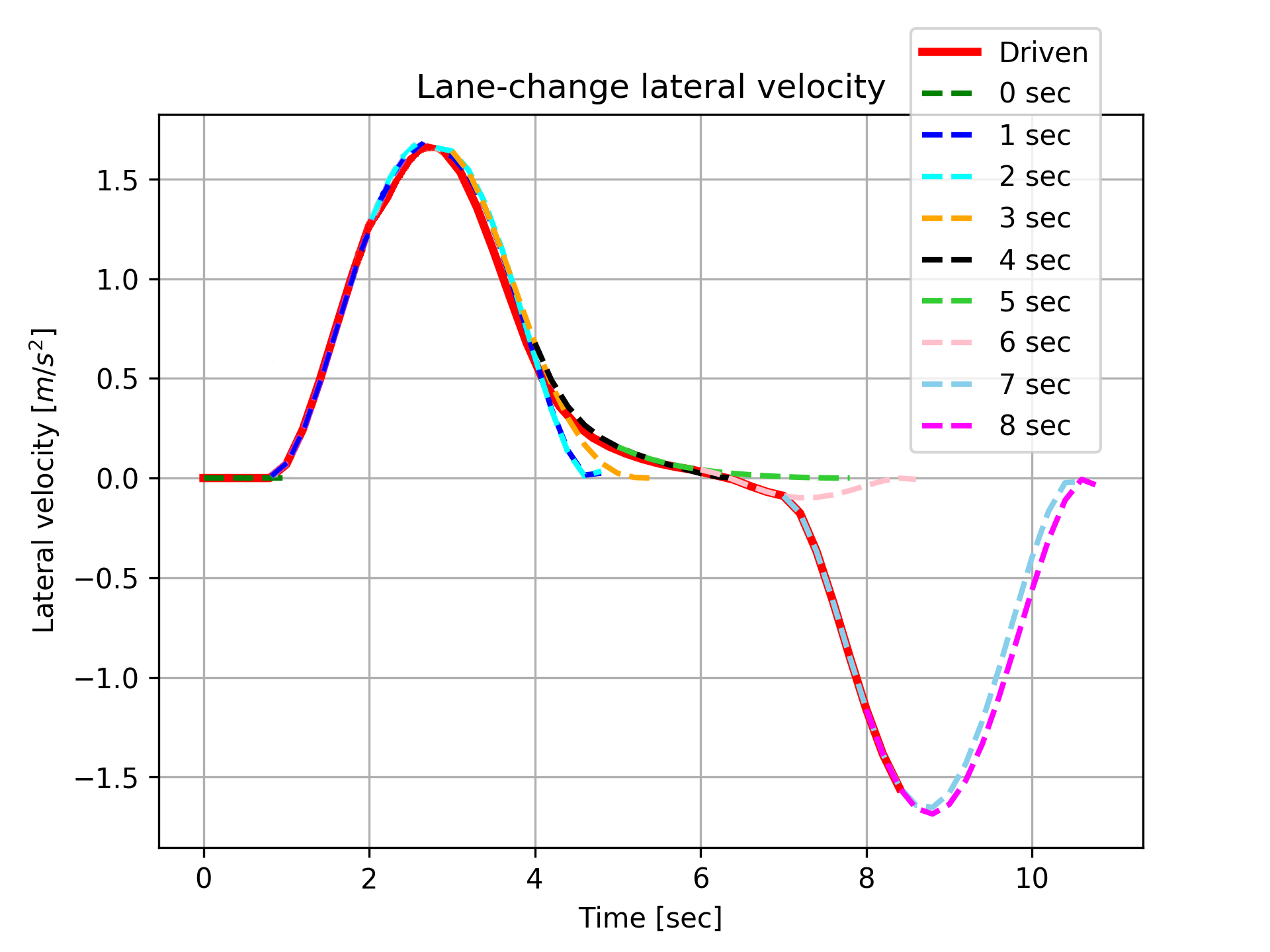}
	\caption{Example trajectory selection during a lane-change. The trajectories chosen each second are indicated with different colors. The complete driven trajectory is the shown in red.}
	\label{lat_velocity}
\end{figure}

The trajectory parameters learning framework consists of four modules: scene understanding, planning/decision making, trajectory generation and trajectory execution, illustrated in Fig.~ \ref{approach_modules}.

\subsection{Scene understanding}
The scene understanding module is in charge of collecting information from the environment and processing it into RL state features relevant for decision making. It consists of information about the RL agent, the surrounding vehicles and the road infrastructure. 
For this purpose we use the DeepSets implementation adapted from \cite{Hgle2019DynamicIF}, which is allowing for variable number of inputs from the environment. 
This is helpful because it alleviates the need to pre-define which and how many objects from the environment are relevant for the decision making.

\subsection{Decision making}
The decision-making module is implemented based on the Twin Delayed Deep Deterministic policy gradient (TD3) algorithm \cite{DBLP:journals/corr/abs-1802-09477}.
TD3 is an actor-critic, off-policy algorithm that is suitable for environments with continuous action spaces.
We randomly initialize an actor (policy) network $\pi_{\theta}$ and three critic networks \footnote{We found that in contrast to Double Q-learning, using three critics improves the performance.} $Q_{w_{i}}, i \in \{1, 2, 3\}$.
Given a fixed batch of training samples $\mathcal{B}$, each training iteration we select a mini-batch of samples $\mathcal{M}$ from $\mathcal{B}$.
In each sample $(s, a, r^{\prime}, s^{\prime}, done)$, $s$ represents the output of the DeepSets, describing the environment in terms of the surrounding vehicles, concatenated with the state-features describing the RL agent.
The complete training algorithm, Offline Trajectory Parameters Learning (OTPL) is described in Algorithm \ref{alg:otpl}.
\indent Once the training is done and we have a well-performing trained agent, we apply it on unseen scenarios for evaluation.
Based on the current RL state $s = (\rho, s_{RL})$, which is a concatenation of the DeepSets processed state of the surrounding vehicles, $\rho$ and the RL agent state $s_{rl}$, the RL agent selects new action, consisting of four continuous sub-actions representing the target trajectory parameters described in \ref{action_space}.
Based on the selected trajectory parameters, a trajectory is generated and checked for safety.
If the trajectory is safe, the first 1 second of it is executed, after which the RL agent selects a new action.
If it is unsafe, the scenario ends unsuccessfully. 
The action execution algorithm for a trained agent is described in Algorithm \ref{alg:trained_otpl}.

\subsection{Trajectory generation} \label{traj_gen}
After the RL agent has made a decision, the chosen continuous actions are forwarded to the trajectory generation module. 
The actions are parameters that describe the desired trajectory. 
Based on the current environment state and on the parameters coming from from the RL agent's action, a polynomial trajectory \cite{5509799} is generated consisting of longitudinal and lateral movement components. 
The longitudinal movement is described using a quartic (fourth order) polynomial, while for generating the lateral movement, quintic (fifth order) polynomials are considered more suitable.
We describe in detail how the trajectory generation works in \ref{appendix}.

Since the RL agent makes a new decision every second, we set the target longitudinal acceleration, target lateral acceleration and the target lateral velocity for each chosen trajectory to zero, without loss of feasibility or smoothness. 
This is because the reward function punishing high longitudinal and lateral jerks makes sure the agent avoids choosing trajectories with short duration and high changes in velocity.
In Fig.~ \ref{lat_velocity} we illustrate an example of a trajectory selection over the course of approximately 8 seconds, in terms of the lateral velocity.

\subsection{Trajectory execution}
After the target trajectory has been generated and checked for safety, it is sent for execution.
Each time-step (0.2 seconds), values from the trajectory for target longitudinal and lateral positions, target velocity and target orientation are sent and executed.
Once five time-steps are done, i.e., after one second, the agent finds itself in the next RL state, where it makes a new decision.


\begin{algorithm}[t]
    \begin{algorithmic}

    \State \textbf{Input}: Random initial parameters $\theta$ for actor policy $\pi_{\theta}$,
	\Statex \hspace{0.89cm} $w_i$ for critics $Q_{w_{i}}, i \in \{1, 2, 3\} $,
    \State \hspace{0.89cm} Fixed replay buffer $\mathcal{B}$, mini-batch size, 
	\State \hspace{0.89cm} Target parameters: ${{\theta}^{'}} \leftarrow {\theta}$; $w^{'}_{i}$ $\leftarrow w_i$,
	\State \hspace{0.89cm} Noise clipping value $c$, policy delay value $d$,  
	\State \hspace{0.89cm} Target policy noise: $\epsilon \sim clip(\mathcal{N}(0, \sigma), -c, c)$,  
	\For{\text{training iteration} $j=1,2,\dots$}\\
		\hspace{0.60cm}{get mini-batch $\mathcal{M}$ from $\mathcal{B}$}
		\ForEach {transition sample $(s, a, r^{\prime}, s^{\prime}, done)$ in $\mathcal{M}$}
		\State {Compute target actions:}
		\State {$\bar{a} \leftarrow clamp(\pi_{\theta}(s^{\prime}) + \epsilon, a_{min}, a_{max})$}
		\State {Compute target:}
		\State {$y \leftarrow r^{\prime} + (1-done)\gamma \displaystyle\min_{1,2,3}Q_{w^{\prime}_{i}}(s^{\prime}, \bar{a})$}
		\State {Update critics by 1 step of gradient descent:}
		\State {$w_i \leftarrow \displaystyle\argmin_{w_i}\frac{1}{|\mathcal{M}|}\sum{(y - Q_{w_{i}}(s, a))^2}$}
		\If{$j \mod d == 0$}
		\State {Update actor by the det. policy gradient:}
		\State {$\theta \leftarrow \nabla_{\theta}\frac{1}{|\mathcal{M}|}\sum_{s\in\mathcal{M}} Q_{w_{1}}(s, \pi_{\theta}(s))$}
		\State {Update targets:}
		\State {$w^{\prime}_{i} \leftarrow \tau w_i + (1-\tau)w^{\prime}_{i}$}
		\State {${\theta}^{\prime} \leftarrow \tau \theta + (1 -\tau){\theta}^{\prime}$}
		\EndIf
		\EndFor
	\EndFor
	\caption{Offline Trajectory Parameters Learning (OTPL)}
	\label{alg:otpl}
    \end{algorithmic}
\end{algorithm}

\begin{algorithm}[t]
    \begin{algorithmic}

    \State \textbf{Input}: Combined state $s = (\rho, s_{rl})$, where 
	\Statex \hspace{0.89cm} $\rho$ is the DeepSets output environment state,
    \Statex \hspace{0.89cm}  $s_{rl}$ is RL agent state, 
	\State \hspace{0.89cm} Trained agent $\pi_{\theta}$
	\While{\text{evaluation scenario not finished}}
		\Statex \hspace{0.60cm}{compute RL action components:}
		\State  \hspace{0.60cm}{$\pi_{\theta}(s) = (a_{tv}, a_{lat_d}, a_{lon_d}, a_{lo})$}
		\Statex  \hspace{0.60cm}{generate trajectory $t$:}
	    \State  \hspace{0.60cm}{$t = generate\_traj(s, a_{tv}, a_{lat_d}, a_{lon_d}, a_{lo})$}
	    \If{$t$ is safe}
	    \State {Execute first second of $t$}
	    \Else \State {$fail$ = True ; break}
	    \EndIf
	\EndWhile
	\If{not $fail$}{ return $success$}
	\EndIf
	\caption{Trained OTPL Agent action selection}
	\label{alg:trained_otpl}
    \end{algorithmic}
\end{algorithm}

\section{MDP Formalization}\label{application}
We consider the problem of safe and smooth driving in realistic driving scenarios, among other traffic participants with different driving styles. 
Additionally, the agent needs to maintain its velocity as close as possible to a desired one.
In this section we describe the RL components and the data used for training.

\subsection{RL state} \label{state}
The RL state consists of two components, one describing the surrounding vehicles in a certain radius around the RL agent, and the other describing the RL agent itself.
For describing the surrounding vehicles, we use:
\begin{itemize}
    \item relative distance $d_{\text{rel}, j}$, defined as $(\text{pl}_{j} - \text{pl}_{\text{rl}})$, where $\text{pl}_{\text{rl}}$ and $\text{pl}_{j}$ are the longitudinal positions of the RL agent and the considered vehicle $j$ respectively,
    \item relative longitudinal velocity $v_{\text{rel}, j}$, defined as $(v_{j} - v_{\text{rl}}) / v_{\text{des}, \text{rl}}$, where $v_{j}$ is the absolute velocity of the vehicle $j$, and $v_{\text{des, rl}}$ is the user-defined desired velocity the RL agent is aiming to achieve,
    \item relative lane $\text{lane}_{\text{rel}, j}$, defined as $\text{lane}_{\text{ind}, j} - \text{lane}_{\text{ind}, rl}$, where ${\text{ind}} \in \{0, 1, 2\}$,
\end{itemize}

The state features describing the RL agent are:
\begin{itemize}
    \item longitudinal velocity, $v_{\text{rl}}\in\mathbb{R}_{\geq0}$,
    \item left lane valid flag, indicating whether there is a lane left of the RL agent, $\text{ll}_{\text{valid}} \in \{0, 1\}$,
    \item analogously for the right lane: $\text{rl}_{\text{valid}} \in \{0, 1\}$,
    \item lateral position $pos_{lat} \in\mathbb{R}$,
    \item longitudinal acceleration, $lon\_a_{\text{rl}}\in\mathbb{R}$
    \item lateral velocity $lat\_v_{\text{rl}}\in\mathbb{R}$,
    \item lateral acceleration  $lat\_a_{\text{rl}}\in\mathbb{R}$.
\end{itemize}
First, the features for each surrounding vehicle will be propagated through the DeepSets module which allows for dealing with a varying number of surrounding vehicles in the decision making process.
Afterwards, it will be combined with the features describing the RL agent to form the final RL state.

\subsection{Actions} \label{action_space}
The agent learns to perform an action $a$, consisting of four continuous sub-actions:
\begin{itemize}
    \item target trajectory longitudinal velocity $a_{tv} \in \mathbb{R}_{\geq0}$,
	\item target trajectory longitudinal profile duration $a_{lon_d} \in \mathbb{R}_{\geq0}$,
	\item target trajectory lateral profile duration $a_{lat_d}  \in \mathbb{R}_{\geq0}$,
	\item target lateral position $a_{lp} \in \mathbb{R}$.
\end{itemize}
These four actions represent the minimal set of parameters needed for generating a trajectory, in compliance with the trajectory generation procedure explained in Fig.~ \ref{traj_gen}.
\subsection{Reward} \label{reward}
The reward function was designed to encourage safe and smooth driving while striving towards the desired velocity.
For a desired velocity $v_{\text{des, rl}}$, given $\delta_\text{vel} = |v^{\text{rl}}_s - v_{\text{des, rl}}|$, the reward function $r: S \times A \rightarrow \mathbb{R}$, is defined as:


Given $sqj_{lon}(a) =  \frac{1}{n}\sum_{i=0}^{n} (j_{lon_i})^2 $, which represents the sum of all squared longitudinal jerk \footnote{Rate of change of acceleration wrt. time} values in the trajectory generated from action $a$, and $n$ is the number of time-steps in a trajectory, a jerk penalty value $jp_{lon}$ and maximum possible jerk value ${j_{lon}^{max}}$ we have:

\begin{algorithmic}
\State $r \gets 0$
\If{$\neg fail$} 
        \If{$v_{rl} >= v_{des}$}
        \State $r += 1 - \delta_\text{vel} / v_{des}$
        \Else
        \State $r += 1$
        \EndIf
        \If{$sqj_{lon}(a) >= {j_{lon}^{max}}$}
        \State $r += jp_{lon}$
        \Else 
        \State $r += jp_{lon}(sqj_{lon}(a) / {j_{lon}^{max}})$
    \EndIf
\Else
       \State $r \gets -0.5$
\EndIf 
\end{algorithmic}

Analogously, we define the same for the lateral jerk: $sqj_{lat} =  \frac{1}{n}\sum_{i=0}^{n} (j_{lat_i})^2 $, $jp_{lat}$, ${j_{lat}^{max}}$ and we add the lateral jerk reward term to the final immediate reward as well. 
$fail=True$ either when the agent collides with other vehicle or leaves the road boundaries.
We provide details regarding the action boundaries and the choice of the jerk penalties in the code repository.
\section{Data collection and Implementation} \label{training_data_impl}
\subsection{Training data} \label{tr_data}
We collected around $5\times 10^5$ data-samples varying the initial position and the behavior of the surrounding vehicles in dynamic highway scenarios.
Our goal is to train the agent on data collected with the most simple policy in an offline fashion.
For that reason we collected the data completely randomly with regards to the action selection.
For the data collection, in order to prevent violations of the maximum acceleration and deceleration, we derived a closed-form solution that given the current state of the RL agent and the chosen target longitudinal duration, outputs a feasible target velocity range.
During execution, the target velocity the agent chooses is capped between the two values calculated by the formula, although a trained agent very rarely chooses a velocity value out of this range. 
We provide more details regarding this formula in the appendix \ref{vel_bound}.
For the other trained agents that we use for comparison in \ref{results}, we also collected the same amount of data with a random policy. 
\subsection{Implementation} \label{implementation}
The parameters used for the Actor and the Critic networks, as well as the rest of the training details are shown in Table \ref{hyperparams}.
For the DeepSets network, we apply the same hyper-parameters as in \cite{Hgle2019DynamicIF}.
The remaining hyper-parameters of the networks are optimized by random search.

\begin{table}[ht]
\caption{Training details}
\begin{center}
\scalebox{1.0}{
\begin{tabular}{l|ll}
    \hline
    & Hyper-parameter & Value \\
    \hline
    Actor &  &  \\
    & Num. neurons per layer & $(400, 300, 1)$ \\
    & Activation functions & $(ReLU, ReLU, TanH)$ \\
    & Policy noise $\epsilon$ & $\mathcal{N}(0, 0.2)$ \\
    & Noise clip & $(-0.5, 0.5)$ \\
    & d & $2$ \\
    Critic & &  \\
    & Num. neurons per layer & $(400, 300, 1)$ \\
    & Activation functions & $(ReLU, ReLU)$ \\
    Other & &  \\
    & $\gamma$ & $0.99$ \\
    & $\tau$ & $10^{-4}$ \\
    & Batch size & $100$ \\
    & Learning rate & $10^{-4}$ \\
    & Max. num. training iterations & $100$K \\
    \hline
\end{tabular}}
\end{center}
\label{hyperparams}
\end{table}

\section{Experiments and Results}\label{results}
To evaluate the performance of our algorithm we designed a set of realistic highway scenarios with a varying number of surrounding vehicles, $N_{veh} = \{10, 20, ..., 80\}$, controlled by a simulator \cite{sumo} policy unknown to the RL agent. 
For each $n_{veh} \in N_{veh}$ there are 10 scenarios leading to a total of 80 evaluation scenarios, where the surrounding vehicles are positioned randomly along the three-lane highway.
In order to capture real highway driving settings as closely as possible, the surrounding vehicles exhibit different driving behaviors within the boundaries of a realistic human driver, i.e., they have different desired velocities, different cooperativeness levels etc.
We let a trained agent navigate through these scenarios and assess its performance in terms of average velocity achieved per set of scenarios for each number of surrounding vehicles. 
\subsection{Comparison to other agents}
\begin{figure}[h]
	\centering
	\includegraphics[scale=0.5]{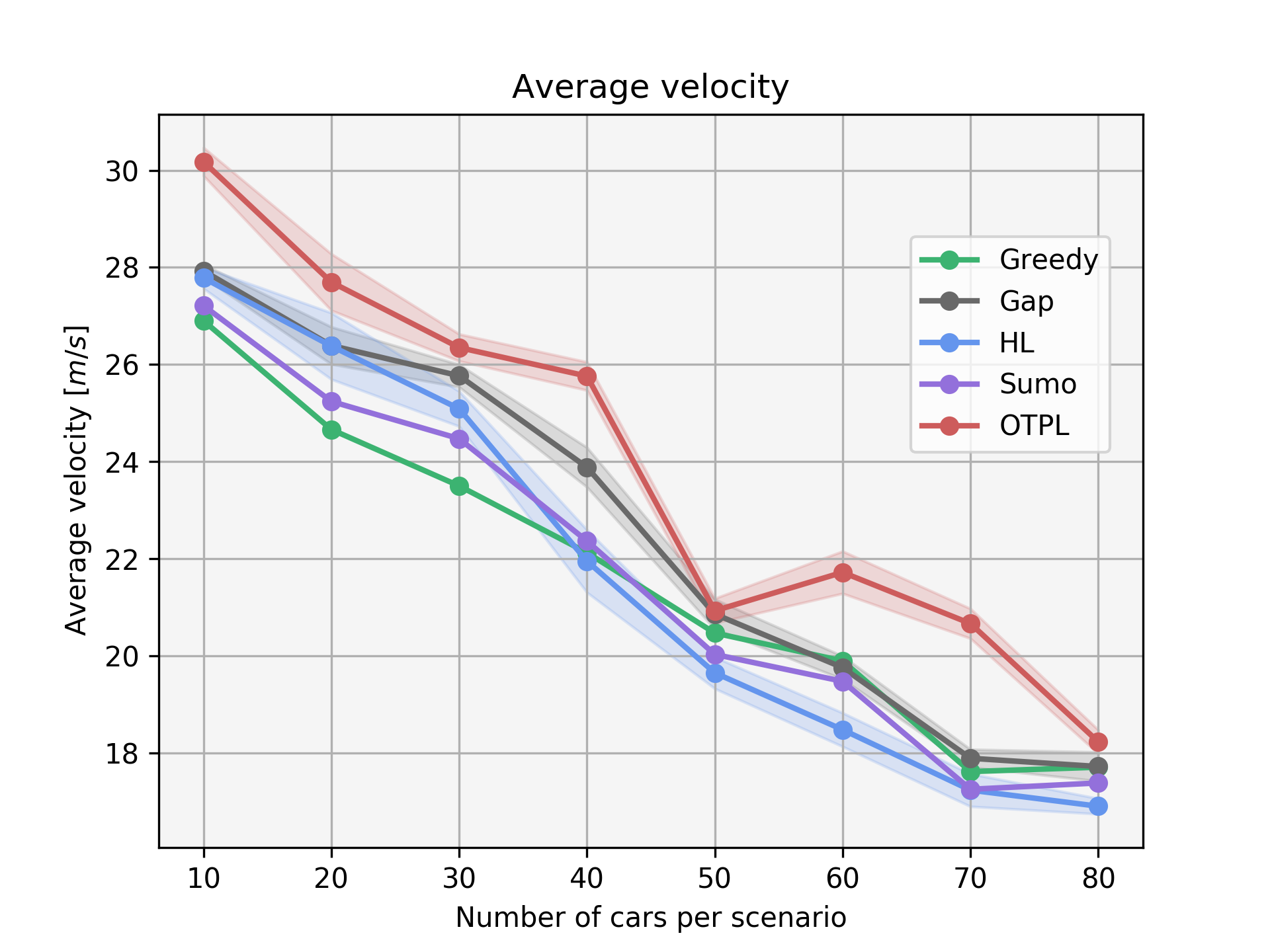}
	\caption{Comparison of the performance of our agent OTPL (red) against the IDM-based (purple), Greedy (green), High-Level (blue) and Gap (grey) agents}
	\label{avg_vel}
\end{figure}

We compared the performance of our algorithm on the 80 scenarios, against four other driving agents. 
An IDM \cite{article} controlled Sumo agent, with the same goals as our agent in terms of desired velocity $(30m/s)$, and parameters concerning driving comfort and aggressiveness levels fit to correspond to the reward function of the RL agent.
The Greedy agent simply selects the gap, to which there is a trajectory achieving the highest velocity. 
The High-Level RL agent is trained to choose from three high-level actions: keep-lane, lane-change to the left or lane-change to the right, and the Gap choosing agent from our previous work \cite{DBLP:journals/corr/abs-2012-03234} trained with RL chooses from a proposed set of actions navigating to each of the available gaps.
The plot in Fig.~ \ref{avg_vel} shows the performance of all agents on the 80 scenarios of varying traffic density.
The x-axis represents the number of vehicles per set of scenarios, whereas the average velocity achieved among each 10 varying-density scenarios is read from the y-axis.
The shown results for the trained agents are an average of the performance of 10 trained agents.
That allows us to show the standard deviation of each of these runs, shown with transparent filled area aground the corresponding curve.
Low standard deviation indicates that there is small variance between models run with different randomly initialized network weights.
Note that after inspection of the scenarios with 50 cars, we determined that there are a few scenarios where the randomly placed surrounding vehicles create a traffic jam and make it difficult for the agents to advance forward. So they are forced to follow the slower leading vehicles, explaining the lower average velocities in that case.
Our RL agent has learned to manoeuvre through the randomly generated realistic highway traffic scenarios without causing collisions and leaving the road boundaries with an average velocity higher than the comparison agents in all traffic densities.

\subsection{Handling critical scenarios}
\begin{figure}[t]
	\centering
	\includegraphics[scale=0.32]{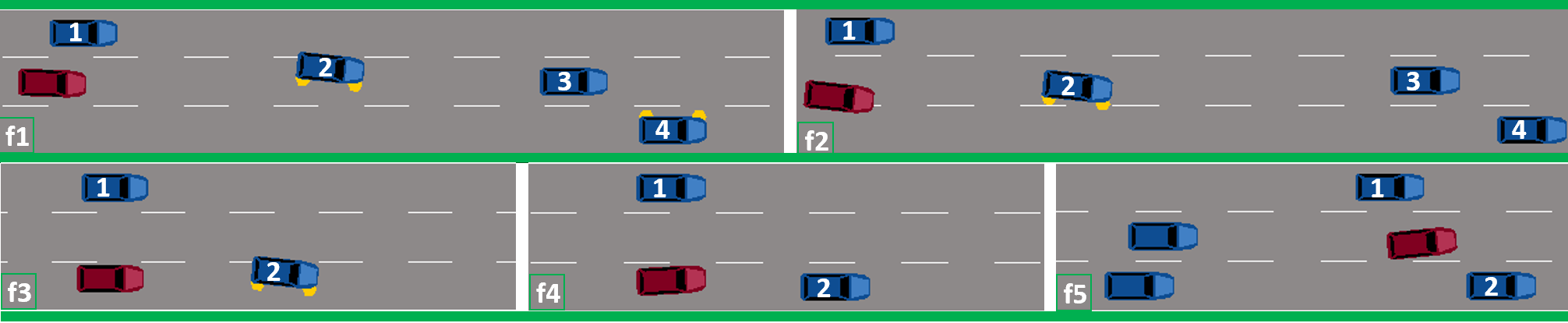}
	\caption{The RL agent (red) successfully avoids collision when the vehicle 2 decides to cut-in in front of it.}
	\label{cutin}
\end{figure}

In order to asses the performance of our agent in critical situations, we created two challenging evaluation scenarios. 
The first one is depicted in Fig.~ \ref{cutin}, where the vehicles with id 2 suddenly decides to change lane twice (scenes f2 and f3), ending up both times in front of our RL agent driving with high velocity.
The RL agent manages to quickly adapt to both situations and avoids collision, finally ending up in the middle free lane, able to advance forward (scene f5).

\begin{figure}[t]
	\centering
	\includegraphics[scale=0.32]{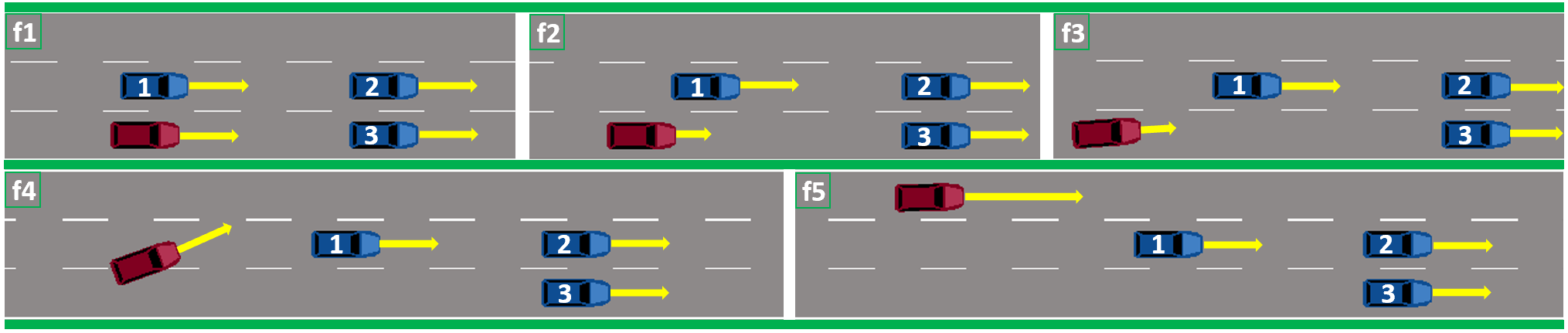}
	\caption{The RL agent (red) learned to perform complex maneuvers consisting of initially slowing down in order to reach the free lane and accelerate towards the desired velocity.}
	\label{trapped}
\end{figure}

In the second scenario in Fig.~\ref{trapped} the RL agent finds itself surrounded by vehicles driving with the same velocity (scene f1), lower than the RL agent's desired velocity. 
This situation might occur for e.g. after the RL agent has entered the highway.
In this case, the RL agent has learned to initially slow down (scene f2) and perform a lane-change maneuver (scene f4) towards the free left-most lane.
This shows that the RL agent has learned to plan long-term, since slowing down is not beneficial with respect to the immediate reward.
The other agents, except the gap-choosing agent were not able to deal with this situation, as they remained in the initial gap until the end.


\section{Training Data Analysis}\label{data_analysis}
Since we are training the agent on a fixed batch of offline data and it does not get to explore the environment based on the current policy, 
the quality of the data is of significant importance.
We performed data analysis to investigate the benefits of the structure imposed by the trajectory polynomials during data collection for the learning progress.
Additionally, we provide analysis of how the percentage of negative training samples affects the learning.
\subsection{Importance of structured data}
A common way of approaching the highway driving task with online RL is teaching an agent to perform low-level actions over small time-step intervals.
Since we are interested in offline RL, in order to evaluate its performance, we randomly collected data with an agent that chooses throttle and steering wheel actions each $0.2$ seconds. 
In addition to other tasks, in this case the agent has to learn not to violate the maximum steering angle, i.e., not to start driving in the opposite direction, as well as not to violate the acceleration constraints.
As a consequence, the collected data consisted of a large portion of such samples, whereas the number of non-terminal and useful samples was negligible.
Additionally, the agent was never able to achieve high velocities since it is interrupted by some of the more likely negative outcomes (like constraint violations) before being able to accelerate enough.
The agent trained with this kind of data did not learn to drive without ending up in an undesired terminal state.
Relying on the structure imposed by the polynomial trajectory generation, helps to obtain high-quality data for offline learning.


\subsection{Percentage of transitions towards a terminal state}
When it comes to terminal states, in our case the agent needs to learn to avoid collisions with other vehicles and to stay within the road boundaries.
We observed that the frequency of such samples in the data and the complexity of the task play a critical role in the performance of the trained agent. Our results show, for instance, that the agent learns to not leave the road very quickly and from much more limited data compared to the more complex collision avoidance task ($4\%$ of data with samples of leaving the road vs. $\sim26\%$ of data with samples of collisions).

\begin{figure}[]
\centering
\hspace*{-0.4cm}\begin{tabular}{cc}
\centering
 a) Average speed & b) Driving time \\
  \imagetop{\includegraphics[scale=0.26]{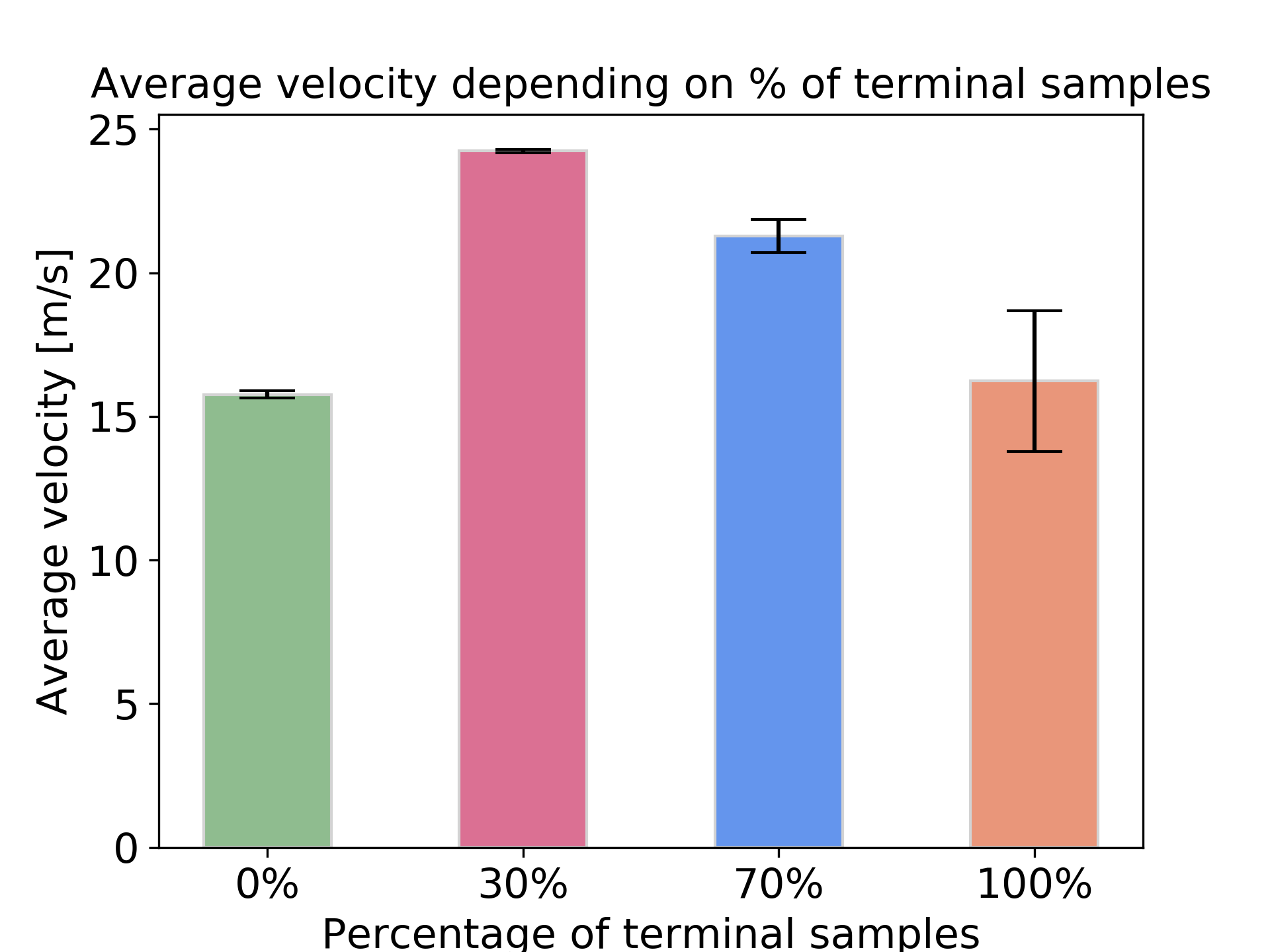}} &
  \imagetop{\includegraphics[scale=0.26]{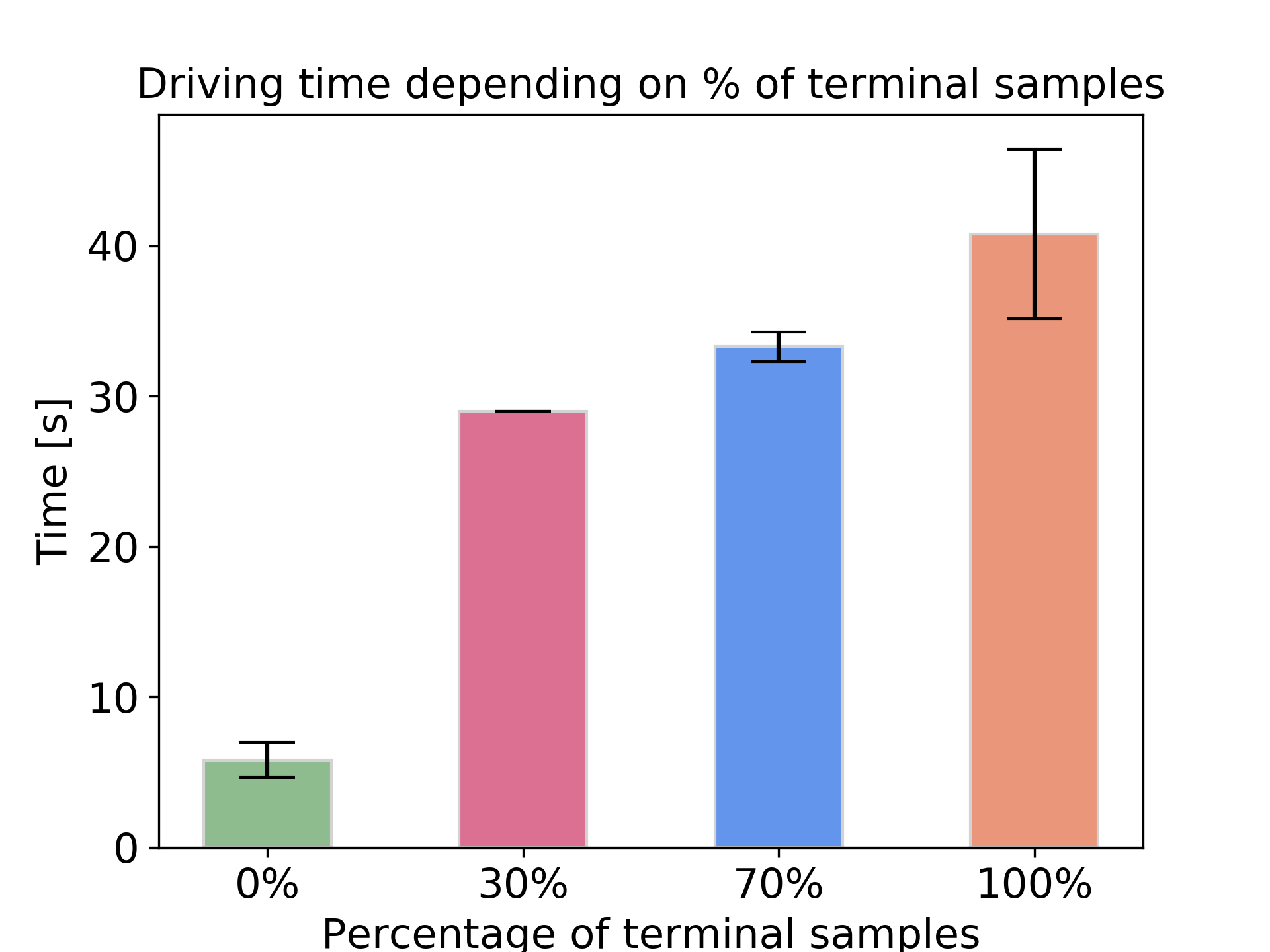}} 
\end{tabular}
\caption{Importance of the percentage of negative samples in the training data. The bars are showing the mean over 5 trained agent for each percentage. The error bars indicate the standard deviation.}
\label{neg_samples}
\vspace*{-0.6cm}
\end{figure}

Fig.~ \ref{neg_samples} a) and b) show the average performance of 5 agents trained with different percentages of terminal samples in the training data.
In a) the y-axis shows the average velocity achieved in the evaluation scenario (randomly generated, 50 surrounding vehicles), depending on the portion of terminal states in the data (x-axis).
Fig.~ \ref{neg_samples} b) represents the driving time of each of the agents. 
Note that the agent trained without any terminal samples never learned to avoid collisions, so it only endures 5 seconds before it collides.
The rest of the agents, learn not to collide and are able to complete the scenario.
The conclusion is, if there are too few terminal samples, the agent never learns to avoid collisions (in the $0\%$ case), since it is not aware of that occurrence.
If there are to many, the agent has problems learning to achieve high velocities, since it has learned to be more conservative.
The best performance is achieved when $\sim 30\%$ of the training samples are terminal.



\section{Conclusion}\label{conclusion}
In this work, we are tackling the problem of teaching an RL agent to drive in realistic, highly stochastic highway scenarios, by learning to optimize trajectory parameters.
Our actor-critic based agent learns to output four continuous actions describing a trajectory that is generated by an underlying polynomial-based trajectory generation module.
Trained only on a fixed-batch of offline data collected with a simple random collection policy, the agent learns to maneuver through randomly generated highway scenarios of variable traffic densities, outperforming the comparison agents wrt. average speed.
On top of learning to avoid causing collisions and to stay within the road boundaries, the agent learns to deal with critical scenarios where more complex maneuvers are required.
Additionally, we provide an analysis of the way the composition of the training data in the offline case influences the performance of the learned RL policy.
\section{Acknowledgement} \label{acknowledgement}
We would like to thank Maria Kalweit, Gabriel Kalweit and Daniel Althoff for the fruitful discussions on offline RL and trajectory planning.
\section{Appendix} \label{appendix}


\subsection{Trajectory generation details}
As mentioned in \ref{traj_gen}, we use quartic and quintic polynomials to describe the longitudinal and the lateral trajectory components, respectively.
More precisely, we need to solve Equations \ref{lon_pos} and \ref{lat_pos} in order to get the positions  wrt. time for the desired trajectory.
In order to do that, we need the values for the coefficients $a_3, a_4, b_3, b_4$ and $b_5$, which considering the parameters given in Table \ref{coefficients}, we can get by solving the systems of equations shown in \ref{lon_poly} and \ref{lat_poly}.
Once we have the positions for the complete duration of the trajectory, we can get the velocities, accelerations and jerks, by calculating first, second and third derivative of the position wrt. the time $t$.
For time-step size we use $dt = 0.2s$.

\begin{equation}
\begin{split}
& traj_{lonp} = a_0 + a_ + a_2t^2 + a_3t^3 + a_4t^4 \\
& \text{where} \quad t = \{0.0, dt, 2dt, ..., a_{lon_{p}}dt\}
\end{split}
\label{lon_pos}
\end{equation}

\begin{equation}
\begin{split}
& traj_{latp} = b_0 + b_1t + b_2t^2 + b_3t^3 + b_4t^4 + b_5t^5 \\
& \text{where} \quad t = \{0.0, dt, 2dt, ..., a_{lat_{p}}dt\}
\end{split}
\label{lat_pos}
\end{equation}

\begin{table}[ht]
\caption{Polynomial parameters}
\begin{center}
\scalebox{1.0}{
\begin{tabular}{lll}
    \hline
    Parameter/Coefficient & Description & \\
    \hline
    $a_0$ & current longitudinal position & \\
    $a_1$ & current longitudinal velocity & \\
    $a_2$ & $\frac{1}{2}$ current longitudinal acceleration &\\
    $acc_{lon}$ & target longitudinal acceleration = 0 &\\ 
    \addlinespace[0.2cm]
    $b_0$ & current lateral position &\\
    $b_1$ & current lateral velocity &\\
    $b_2$ & $\frac{1}{2}$ current lateral acceleration &\\
    $vel_{lat}$ & target lateral velocity = 0 &\\
    $acc_{lat}$ & target lateral acceleration = 0 &\\
    \hline
\end{tabular}}
\end{center}
\label{coefficients}
\end{table}

\begin{equation}
\begin{cases}
a_1 + 2a_2 + 3a_{lon_{d}}^2a_3 + 4a_{lon_{d}}^3a_4 + a_{tv} = 0\\
2a_2 - acc_{lon} + 6a_{lon_{d}}a_3 + 12a_{lon_{d}}^2a_4 = 0
\end{cases}
\label{lon_poly}
\end{equation}

\begin{equation}
\begin{cases}
\text{\footnotesize $b_0 + b_1a_{lat_{d}} + b_2a_{lat_{d}}^2 + b_3a_{lat_{d}}^3 + b_4a_{lat_{d}}^4 + b_5a_{lat_{d}} - a_{lp} = 0$}\\
\text{\footnotesize $b_1 + 2b_2a_{lat_{d}} + 3b_3a_{lat_{d}}^2 + 4b_4a_{lat_{d}}^3 + 5b_5a_{lat_{d}}^4 - vel_{lat} = 0$}\\
\text{\footnotesize $2b_2 + 3b_3a_{lat_{d}} + 12b_4a_{lat_{d}}^2 + 20b_5a_{lat_{d}}^3 - acc_{lat} = 0$}
\end{cases}
\label{lat_poly}
\end{equation} 

\subsection{Action target velocity boundaries} \label{vel_bound}
As mentioned in \ref{tr_data}, we derive a formula based on which, given the current state and $a_{lon_{d}}$ we calculate the minimum and maximum target velocity values the agent is able to choose from.
This is done so that the generated trajectory will not violate the minimum/maximum possible longitudinal acceleration.
This contributes for collection of data that consists of informative samples that enable fast learning of a good policy, instead of data-samples that are repetitive and lack useful information.

Additionally, this way we avoid learning the minimum/maximum acceleration, which simplifies the learning process.
By calculating the second derivative of Eq.~ \ref{acc_t}, we get the trajectory acceleration equation.
\begin{equation}
acc_{lon}(t) = 12a_4t^2 + 6a_3t + 2a_2
\label{acc_t}
\end{equation}
By calculating the first derivative of the acceleration equation and setting it to 0, we get the expression for obtaining its minimum/maximum.
\begin{equation}
24a_4t + 6a_3 = 0
\label{acc_t}
\end{equation}
After solving for $t$ we get the time-step at which the minimum/maximum acceleration is achieved.
We substitute the expression for $t$ back to the acceleration Eq.~ \ref{acc_t} and we set it to the known minimum/maximum acceleration value.
After finding the solutions for $a_4$ and $a_5$, we can substitute them in Eq.~ \ref{lon_poly} and express the values for the minimum and maximum $a_{tv}$ allowed.
We use \cite{10.7717/peerj-cs.103} to solve the equation and get the final values for the minimum and the maximum allowed velocities.
We will provide the script in our code repository \href{https://nrgit.informatik.uni-freiburg.de/branka.mirchevska/offline-rl-tp}{here}.

\printbibliography
\end{document}